# Dynamic and Systematic Survey of Deep Learning Approaches for Driving Behavior Analysis

Farid Talebloo, Emad A. Mohammed, Behrouz H. Far

Department of Electrical and Software Engineering

[farid.talebloo, eamohamm, far] @ucalgary.ca

University of Calgary, 2500 University Drive NW, Calgary Alberta T2N 1N4, CANADA[1]

## Abstract

Improper driving results in fatalities, damages, increased energy consumptions, and depreciation of the vehicles. Analyzing driving behaviour could lead to optimize and avoid mentioned issues. By identifying the type of driving and mapping them to the consequences of that type of driving, we can get a model to prevent them. In this regard, we try to create a dynamic survey paper to review and present driving behaviour survey data for future researchers in our research. By analyzing 58 articles, we attempt to classify standard methods and provide a framework for future articles to be examined and studied in different dashboards and updated about trends.

## Keywords

driving behaviour identification, driving behaviour analysis, dynamic survey, deep learning approaches, intelligent transportation systems.

## Abbreviations

DBA = Driving Behaviour Analysis

RNN = Recurrent Neural Network

LSTM= Long Short-Term Memory

WHO= World Health Organization

GPU= Graphic Processing Unit

AE= Auto Encoder

IEEE= Institute of Electrical and Electronics Engineers

ACM= Association for Computing Machinery

DL = Deep Learning

ML = Machine Learning

AI = Artificial intelligence

CNN = Convolutional Neural Networks

RNN = Recurrent Neural Networks

## Introduction

The lives of nearly 1.35 million people are lost every year due to road traffic accidents [1]. Between 20 and 50 million more people suffer non-fatal injuries, many of whom experience disability due to injuries [1]. More than 32,000 fatalities occurred in the US in 2013 due to drunk driving and speeding, leading to more than 19,500 deaths [2]. Even with these staggering numbers in the US, China is the nation with the most road traffic accidents in the world. There were 8934 traffic accidents in 2016, causing 5947 deaths and 11,956 injuries on China's freeways. Although freeways account for just 2.8% of the overall length of public roads in China, traffic collisions, casualties and fatalities accounted for 7.7%, 9.4% and 13.7% of all traffic accidents, injuries, and deaths. Compared to other road grades, the collision rate, damage rate and death rate per 100 km of Chinese highways are 3.0 times, 3.8 times and 5.1 times higher, respectively [3].

[1] Acknowledgements: This research is partially supported by NSERC Discovery Grant and Alberta Major Innovation Fund (MIF).

One of the innovations developed in vehicles today to meet the need for protection and comfort in driving is Advanced Driver Assistance Systems or ADAS. ADAS are vehicle control systems that use environmental sensors (such as radar, laser, vision) to enhance driving comfort and safety by helping drivers identify and respond to potentially unsafe traffic situations. One of the primary ADAS studies relates to model driving behaviour. Driving behaviour research needs to be carried out because, based on WHO data, driver factors are among the leading causes of vehicle accidents. Examples of driver factors include over-speed speeding, drunkenness; exhaustion; dark road driving; impaired visibility; and vehicle quality factors [4]. The most vital aspect of on-road driving safety is the behaviour of drivers. The level of driving skills (e.g., professional; beginner) affect driving behaviour, and the set of decisions made at any given moment in driving may lead to one type of driving behaviour at each stage. On the other hand, driving events such as acceleration, deceleration; turning; braking lead to driving behaviour. These two separate sets are affected by the prevailing driving conditions such as traffic, weather, cars, and roads [5].

## Scope of research

The scope of this survey paper includes all the research work performed between October 2015 to September 2020, having keywords with "Driving Behaviour Analysis" and "DL algorithms," focused on computer-related journals and databases includes IEEE, ACM, Scopus, and Springer. We exclude traditional ML, psychological analysis of driving, and any paper out of the mentioned dates. Moreover, fortunately, with the algorithm of the "dynamic survey," which means "adding papers dynamically in future," our database will include all related papers on future dates.

## Dynamic Survey Approach

It is common for "Survey papers" to usually limit the scope and study them so that the researchers will review them over a period. Consequently, after publishing the paper, any activity such as new articles or new chapters of a book-related to the topic to be published after the survey paper should be considered in a new survey paper. Nevertheless, in this dynamic survey, the included articles will be updated using an algorithm in its database.

In the following, we explain how the algorithm works: in this mechanism, first, we adjust the recurrence period (i.e., daily, weekly, monthly) of the web requests. Then, it starts sending web requests to the APIs of defined journals (i.e., IEEE, Elsevier, Springer, …) to get the list of newly submitted articles in DBA. After receiving the list of newly published articles, it compares them with the existing list in the database to avoid duplicating an article.

Finding a new paper submitted related to DBA is not enough; we needed a space to store the data. To keep the data normalized, we decided to use a relational database (i.e., MS SQL Server). Other advantages of using the relational database are avoiding redundant data, ease of inquiry, easily transferable, high security, and high efficiency. In designing a normalized database, we tried to make it article-centric. The newly found article title and core information will store in the main table, and other pieces of data will be stored and pointed to it [Figure 2].

In addition, we develop a dashboard to display all articles' values as an interactive chart in Google Data Studio [6]. The researchers can use these charts to find out the trends of driving behaviour analysis in the future. They can change the dimensions, change the sorting types, zoom in and out in geographical charts to study this topic further [Figure 3]. In [Figure 3], there are four sortable charts that we will explain their functionality of in the following: 1) Sorted the countries that have more studied paper 2) Sorted the approaches that research have been used to tackle DBA studies 3) Sorted keywords of studied documents in this study 4) Showed the trend of paper counts from 2015 to 2020

[Figure 4] shows the geographical distribution diagrams of studies conducted in the field of driver behaviour analysis. In the figures above, a data set is presented in two different forms to make the study more accessible. However, in the table below, the information related to the approaches used by the researchers and the country in which the research was conducted is shown appears as a matrix.

In [Figure 5], we used a treemap diagram to look at the data set from four angles. The chart on the top left of the scale shows the number of times researchers used the solutions. The chart on the top right shows the scale of the number of surveys in countries. The chart on the bottom left shows the scale of the number of times keywords are used in papers. Furthermore, the right bottom diagram shows the number of attributes of each article. In [Figure 6], we develop a dynamic sortable table that includes approach and citation counts of each approach used by researchers.

All the developed files and processes are added to the GitHub repository [7]. Our database design mandates adding a newly published paper in the driving vehicle behaviour as the main object of a referable record. All the other values can be attached to the primary node (Name of the article). Moreover, concepts such as authors, affiliations; type of input and output data; and result presentation may be added.

# Background of the Driving Behaviour Analysis

## Declarations of Driving behaviour Analysis

It is helpful to define Driving Behaviour Analysis (DBA) by considering the term word by word. Analysis can be said to refer to *"The process of studying or examining something in an organized way to learn more about it or a particular study of something."* [8]. *Behaviour* is typically understood as" the response of an individual, group, or species to its environment" [9]. Finally, driving can be defined as" to operate the mechanism and controls and direct the course of (a vehicle)" [10]. Taken together, then, the meaning of DBA is to detect and study drivers' behaviour by leveraging the output data gathered from their vehicles. Drivers can be identified for unsafe driving behaviour such as harsh accelerating, sharp slowing; frequent braking; speeding; harsh high-speed turning; sluggish driving, frequent parking, and fatigued driving [IBM]. Driving behaviour insights can help automotive manufacturers improve design and manufacturing, assist with quality control; enhance protection, and simplify maintenance. This can help manage vehicle fleet activities, and insurance providers can get powerful insights into the vehicle use and risk evaluation of their customers. Considering the importance of DBA from another angle, we find that one of the most critical consumers of this algorithm is insurance companies that can take advantage of dynamic insurance systems by ensuring that each driver will be charged according to their driving behaviour.

## History of Driving Behaviour Analysis

The first car built and offered for purchase occurred in the early twentieth century. Consequently, at that time, all efforts were focused on car mechanics [11]. The "Duesenburg Model A" became the first vehicle to have hydraulic four-wheel brakes in 1922; this is considered the first step in increasing car safety [12]. One of the first studies to analyze driver behaviour was the paper by Hsing-Shenq Hsieh et al. [13]. They assumed that most video systems were ineffective at unsaturated intersections due to the chaotic combinations of driver behaviour and the geometric design of curved lines. They defined their methodology using several reference points, and the Bleyl transfer method was developed as a matrix and the least-squares method (LSM).

## Deep Learning Methods

AI aims to give computers the mental power to program them to learn and solve problems. Its purpose is to simulate computers with human intelligence [14]. In recent years, various algorithms in the analysis and understanding of driving behaviour have been used. To the best of our knowledge, we explore the papers scoped in the "scope of research"; related to the DBA concept. Furthermore, we explore the DL approaches those researchers used to prepare results. In the next step, we have investigated the citation number of each paper and tried to find out the best of them regarding their method and approach popularity.

Our paper focuses on studies that used DL for this research. Instead of defining the hierarchy and layering of existing algorithms classically, we address the popularity of algorithms to provide a background for analyzing and categorizing existing articles in this field. The following section lists the most popular methods to detect driving behaviour patterns in the researcher's papers.

### Convolutional Neural Network

The term CNN means that a statistical operation called Convolution (a specialized sort of linear reaction) is used for the network [15]. Convolution is a function derived from two functions by integration (in strictly mathematical words), which describes how one's form is changed by another [16]. The most significant benefit of CNN relative to its contemporaries is that it automatically identifies the vital features without any human supervision [17]. CNN has several critical drawbacks: many training samples are needed for learning weight parameters, and a strong GPU is needed to speed up the learning process. Also, the strong CNN is often not taken advantage of by researchers who do not have such computing resources, time, and large-scale training data [18]. In the reviewed literature, the most popular methodology is the CNN algorithm.

### Recurrent Neural Network – Long Short-Term Memory

A subset of neural networks used to analyze sequential data input or output are RNN. By providing feedback to Feedforward neural networks, RNNs have a temporal relationship between input/output sequences [19]. RNNs use the previous outputs as inputs that have hidden states. The advantages of RNNs are 1) the possibility to process the input of various lengths, 2) the fact that the size of the model does not increase with the input size; and 3) the reality that the history of the previous information is included in the calculation, and that the weights are shared over time. However, RNN has several disadvantages, including 1) slow calculation, 2) difficulty accessing information from long-term history, and 3) that future inputs cannot be considered for the current situation [20].

### Auto Encoders

AEs are practical learning approaches that aim to transform input to output with the least possible error [21]. An AE finds a representation or code to classify different inputs to perform valuable transformations on the input data. For instance, a neural network finds a code that can turn noisy data into clean data by denoising AEs. Noisy data can be transformed into coherent sound in an audio recording of static noise [22].

AEs require considerable computational resources and data. The functionality we are attempting to retrieve can often be blurred by a loosely organized testing dataset used in AE training. AEs can be coupled with various neural network architectures, such as feedforward NNs, CNNs, and RNNs, in semi-supervised learning activities. The combinations listed previously can provide good results in multi-task ML problems by simplifying the inputs to a representative code. Nevertheless, it can also harm the model's interpretability even more as it is a trade-off for researchers to select between the simplicity of network or complexity of interpretation. AEs can collaborate with other neural networks independently in unsupervised and semi-supervised learning activities, considering their drawbacks [23].

### Self-Organizing Map

The Self-Organizing Map (SOM) projects a high dimensional distribution on a primary grid that is in order. SOM converts complicated nonlinear relationships between high-dimensional data objects to more straightforward geometric relationships. Compressing the most important topological and metric relationships while preserving specifics will result in complex abstractions. SOM process analysis, perceptions, and communications in complex activities [24].

## Key-Papers Review

In this section, we review the most cited papers related to our review. A more detailed summary is illustrated in Appendix A. Recent work by Weishan Dong et al. [25] examined and identified driving behaviour, extracted them, and added five new statistical features using only latitude and longitude characteristics; it was essential for them to to help the model interpretation. The authors used GPS data only and claimed that in the future, with complete information, such as On-Board Diagnostics (OBD) data and other sensors, the results will be more accurate. They then split the data into specific frames. Each frame was labelled "Driver Id," and the method was "supervised." First, the researchers used the method of Yang et al. [26], followed by the use of the Integrated Recurrent Neural Network (IRNN) method, plus the "two stacked recurrent layer." In addition to these two methods, Dong et al. utilized "non-deep learning" methods. Gradient Boosted Decision Tree (GBDT) and "TripGBDT" methods employed the Kaggle site dataset for implementation; In this method, statistical data is used along with the available features (57 in total). Several studies investigated whether the deep learning methods (i.e., hidden global travel information) perform better than the GBDT method (i.e., information is shared as a feature). One result showed that if the sampling rate were less than 0.1 Hz (a record in 10 seconds), the results would be severely low. The Kaggle 2015 competition on driver telematics analysis data [27] was used to test the proposed researches. Dong and colleagues found that the derived traits were not as strong as those learned by DL.

Furthermore, the "Stacked-IRNN" method was seen to perform better than the others. However, it naturally costs a lot to converge over time. The researchers noted the operational issues that DL methods for online prediction are much more helpful than "TripGBDT." Privacy is one of the critical issues in the use of telematics data. Furthermore, road shape, traffic, and weather can affect driving behaviour. Also, vehicle sensor data such as OBD can be added to the model as a feature.

Pengyang Wang et al. [28] assumed that driving activity is a dynamic task that involves professional multi-level movements, such as acceleration, deceleration, constant speeding, left turning, right turning, and straight movement that would be complex interpretation for an AI model. Wang et al. claimed that studying driving behaviour will help to analyze driver performance, improve traffic safety, and eventually facilitate the development of intelligent and resilient transport systems to allow many critical applications, such as tracking drivers, automobiles, and highways; providing early warning and driving assistance; improving driving comfort; and saving energy. Three distinct types of DBAs exist: 1) Descriptive analysis is the first type, in which transport experts identify metrics (e.g. harsh or repeated acceleration/braking, sharp turn, or pre-turn acceleration) based on a transport theory that explains driving behaviour [29]. 2) In the second type, predictive analysis, researchers utilizing driving data patterns and ML methods (e.g. SVM; Naïve Bayesian) to forecast vulnerable scores [30]. 3) Casual analysis describes the causal factors in driving behaviour and demonstrates how these factors affect road safety [31].

Wang et al. developed a peer and temporal-conscious representation learning-based analytical framework for DBA using GPS tracks. Firstly, a series of Multiview driving state change graphs from GPS tracks were created to describe each vehicle's complex driving activity. Secondly, graph-graph peer and current-past time-dependent driving behaviour patterns were defined, and peer and time-dependent modelling were integrated into a single auto-encoder-based optimization system. Driving behaviour representations for estimation and historical evaluation were studied, with risky zone identification as implementations. Finally, detailed tests were performed to demonstrate the proposed system's improved efficiency with GPS tracks in real-world cars. The researchers described two phrasal terminologies that allowed them to incorporate their method. *Driving Operations* are defined as a collection of actions and measures that a driver operates while driving a car, according to the driver's judgement, expertise, and skills. *Driving State* is concerned with how vehicles travel at a certain time point or in a short time window. In other words, a driving condition of a car varies with time, which involves the speed status (i.e., acceleration, deceleration, steady speed), and the path status (i.e., turning left, turning right, going straight). A series of driving states may be: [<acceleration, moving straight>, <constant speed, moving straight>, <deceleration, turning right>]. The suggested approach took advantage of Multiview driving state transfer graphs. Different observers interpret the transition from two different perspectives: the likelihood of transition and how long the transfer continues.

Three objectives were followed in the model design. First, structural consideration: by designing the desired graphs, they were transformed into vector data. A second objective was peer dependency: Drivers who mimic each other's actions, patterns, and attitudes. The model should reflect them according to the graph-graph peer dependency model. The third objective was temporal dependency: The current time slot's driving operations demonstrate clear autocorrelation connected with the previous driving states.

Wang et al. [28] analyzed driving habits from the context of representation learning. They considered how fast and how long people travelled by building driving state transition graphs. They investigated how one specific driving behaviour relies on another driving behaviour. The framework was generated by empirical modelling of the interconnections between the peer and temporal dependencies. They first defined driving behaviour using Multiview driving state graphs. They developed the idea of graph-graph (definition = trajectories with similar driving behaviour in the graph-graph peer dependency should have near representations in the learnt representation feature space [28])peer penalties to capture the temporal dependency of a single G-G peer by contrasting a graph-graph peer's present value with its initial value. They also applied the device to detect hazardous routes and rank drivers according to driving behaviour automatically. Test runs on real-world data have shown that Spatiotemporal Representation Learning is efficient for driving.

Jun Zhang et al. [32] recommend a DL system for behaviour identification by fusing convolutional and recurrent neural networks, called attention-based DeepConvGRU (Convolutional, Gated Recurrent Unit) DeepConvLSTM. First, in-vehicle sensor data via CAN-BUS is gathered to classify drivers' driving habits. The data was separated into parts for the method of normalization and sliding window study. Finally, the derived driver action patterns were used in a "deep learning" algorithm for recognition. Their key contributions are described as follows: their architecture conducted automated behaviour detection on real-time multi-dimensional in-vehicle "CAN-BUS" sensor data, capturing local dependence among the temporal component (i.e., velocity, acceleration) data and across spatial locations. By incorporating the attention function, their model may catch salient structures of high-dimensional sensor data and explore the correlations among multiple sensor data channels for rich feature representations, enhancing the model's learning efficiency. Their architecture can be used to train end-to-end with no function engineering (which means, instead of adding statistical functions to add more value to the dataset), utilizing raw sensor data without preprocessing relevant to any sort of sensor.

The GRU/LSTM cells distinguish driving habits by adding historical habit values to temporal values. The "Deep Convoluted GRU" model used DL to exploit temporal dynamics. The proposed approach outperformed the conventional system on the "Ocslab driving dataset[33]". The proposed methodology learnt features from the original signals and fused the learned features without any special preprocessing. Surprisingly, the DeepConvGRU obtained competitive "F1 ratings" (0.984 and 0.970, respectively) using at least 51 raw sensor input channels.

Jooyoung Lee et al. [34] established a method to analyze in-vehicle driving data and demonstrate possible violent driving signs. The mentioned system for detecting sudden shifts was based on a two-tier clustering strategy. Some researchers have utilized these methods to detect sudden shifts in driving and cluster driving incidents. With this process, actual in-vehicle driving records of taxis in Korean metropolitan cities were examined. The clusters were used to assess whether another driver's driving record is a possible risk of violent driving and include statistics on potentially aggressive driving.

The research of Jooyoung Lee et al. [34] was performed sequentially to recognize violent driving habits. The technique comprised a three-stage advancement of a mission, detecting a transition, and extracting functions. An in-vehicle recorder was developed to document driving conditions over time. As the data on RPM (revolutions per minute), acceleration and yaw rate are used, the model's precision improves as the three time-series data display significant improvements. When the change point is observed, they decided it has passed the 5th percentile of the results. Once a rapid shift in direction, acceleration, and yaw rate are observed, they can identify the phenomenon as a driving event. Using an unsupervised learning process, the researchers gathered sudden shift events (unexpected shifts) and categorized instances (driving incidents). The framework can evaluate driver behaviour and give recommendations to drivers on their driving style. They think it can be a helpful tool for driver education because driving records may include driving behaviour in real road conditions, unobservable in controlled environments. Although the driver's self-reported aggressive driving incidents could be related to collisions, it is not yet confirmed if aggressive driving events or accidents are connected. The authors suggested they might have potentially affected their study's findings through driver characteristics, but driving reports have already collected these factors. Researchers could not analyze the effect of other variables on driving activity because personal data and sensor limitations hindered precise estimates of the other factors.

## Reviewing the datasets

The "Kaggle. Driver Telematics Analysis" [27] dataset has a directory containing multiple files. There are 200 CSV files found inside each folder. Each file defines a driving trip. The trips are recordings of the location of the vehicle per second (in meters). The trips were based on starting at the origin (0,0), arbitrarily rotated, and short lengths of trip data were omitted from the start/end of the trip to safeguard the privacy of the drivers' locations. A small and random number of false trips (trips that the driver of interest did not drive) are put in each driver's folder. The number of false trips or a labelled training set of real festive trips is not given (it varies). In any given folder, most of the trips belong to the same driver [27].

Driver ID, order ID, time, latitude, and longitude are included in the "Didi Chuxing GAIA Initiative to the research community" dataset. The GPS trajectory's precision is 3 s, and the tying lane processes it. With picking the one-month drip taxi data for October 2016 in Xi'an, China; A more straightforward scoring system was used to mark this dataset [26].

The "UAH-DriveSet" is a public data array captured in multiple settings by the driving tracking software "DriveSafe," by separate testers, supplying a vast number of recorded and processed variables across all smartphone sensors and capabilities during independent driving experiments. The application was tested on six different drivers and cars, with three different behaviours (normal, drowsy, and aggressive) performed on two types of roads (motorway and secondary roads), resulting in more than 500 minutes of naturalistic driving with its related raw data and additional semantic knowledge, along with video records of the trips [35].

It was legislated in Korea in 2011 that all commercial vehicles (e.g., cars, buses, and taxis) must have a digital tachograph (DTG) fitted, a kind of in-vehicle driving recorder for safety monitoring, to capture the "DTG database" dataset (Traffic Safety Act Article 55 in Korea). The DTG device imports the driving documentation from the On-Board Diagnostic Systems (OBD-II) terminal and stores them on the Secure Digital memory card (SD). Data stored in the memory is periodically extracted and transmitted to the Korea Transport Protection Authority (TS) server via the internet [34].

"OSeven Telematics, London, United Kingdom": Data is obtained from an already established mobile program on both iPhone and Android smartphones. The program is still running in the mobile operating system such that no user intervention is taken when commuting. The program gathers raw data from smartphones using multiple parameters using accelerometers, gyroscopes, and GPS cameras. In m/s2, the accelerometer will record the acceleration of a smartphone in terms of gravity acceleration, while the gyroscope measures the angular velocity of the smartphone in rad/sec. Finally, GPS data is obtained to monitor the speed of the vehicle and the vehicle's coordinates. Because the program uses cloud-based services, data is transmitted to the server for storage anonymously for further analysis after automated identification at the end of the ride [36].

The "KITTI" Data Collection includes the specifics of raw gray stereo squares; natural colour stereos and colour squares; 3D Velodyne point clouds; 3D GPS/IMU data; calibrations, and 3D entity track-list marks, which can then be processed and registered at 10 Hz. Directories and directories with dates of formation are presented. Most consumers need to transform and cleanse the data after processing it [2]. KITTI dataset indicates a one-week route of 10.357 taxis in the "T-Drive trajectory dataset." This dataset comprises approximately 15 million points and a cumulative distance of nine million kilometres [37].

The HCRL dataset comprises a ride from Korea University to the SANGAM World Cup Stadium of ten drivers, a cumulative driving period of around 23 hours, and a route that involves riding via Seoul and the surrounding areas for about 46km [33].

The "Warrigal" dataset is a broad, rich dataset extracted in an industrial environment from the experiences of large trucks and smaller 4WD vehicles. A fleet of 13 vehicles working in a surface mine for three years collected the results. Information about the vehicles' status (e.g., location, speed, and heading) and their peer-to-peer radio contact descriptions are contained in the dataset. With a resolution of 1 Hertz, the data extends three years. To the best of our knowledge, no other publicly accessible data collection comes near this degree of information over such a significant period. There is no precedent for the research possibilities and applications that these data allow. This dataset has already been used to analyze map formation, protection analysis, driver purpose inference and wireless network antenna failure [38].

## Survey Dimensions

After reviewing the articles published in this field, we categorized them by the method used for DL, concluding that CNN, LSTM and AEs were the most common methods [Figure 7]. In terms of publications, we have categorized and sorted the articles. The largest share of article publishing was found to be in Scopus, IEEE, and ACM, respectively [Figure 8]. We have found that some researchers do not make available their research data, which is typically owned by a specific organization and only report the results of their research. At the same time, some other researchers made their data available to the public. The proportion of these two types in the total number of articles we considered is 37 to 19, respectively [Figure 9]. There are times in the research process when a researcher uses several datasets to prove a proposed theory or algorithm. Examining the available articles, we found that the number using more than one dataset is seven, versus 33, which used one input database.

## Future Direction and Outlook

Following the publication of this paper, we decided to enhance the accuracy of driving behaviour identification using deep learning neural networks and the "LSTM" technique under the "RNN" branch. Using one of the available datasets, we attempt to build, test, and assess the required modelling.

Following that, we will expand on the technique we presented to implement survey papers so that researchers may use it as a general open-source library for any research publications, independent of the subject covered in this study.

## Challenges and Opportunities

We report the challenges of researchers in this field and then address our challenges in this review. One of the most critical challenges facing researchers in this field is finding suitable datasets to perform various calculations to improve their algorithms. Usually, data heterogeneity presents another challenge once data sets are found, meaning that the measurements made during the data collection process may not have been carried out at the same time interval. Researchers need to homogenize their data at this stage. Next, researchers must decide what features to accept in their proposed model.

Furthermore, because the scope of our review lies in the field of DL algorithms, researchers have tried to use all the features and sometimes even add the calculated features to them. For example, instantaneous velocity was calculated and sent to the model by having different points in the associated times. Some researchers question these calculated features and believe that the deep ML model extracts them.

A challenge related to our review is the number of papers to be surveyed and the adoption of inclusion and exclusion criteria. Several publications took place over different periods and in separate databases. The lack of a centralized database update motivates us to redesign our survey methodology to be dynamic to cover future work in this field. We further implement a dynamic relational database management system that periodically crawls several pre-set databases. Moreover, the dynamic database design allows other databases under user control with different inclusion/exclusion keywords. Another challenge to our survey methodology is the means of actively representing extracted information from reviewed papers. Most of the previous work provides static views of this information that may not reflect the future directions in this field. We propose actively demonstrating the extracted information using an interactive platform (i.e., Google Data Studio). In addition to the challenges mentioned above, another issue we encounter when developing this dynamic database is best communicating with different interfaces in different database search engines. We further address the inconsistency in database interface by implementing an adaptable interface in dynamic database design that can easily connect to several interfaces.

## Conclusion

Accidents impose severe socio-economic complications for the community, and thus safe driving behaviour is a critical component in saving lives on the roads. Considerable research efforts were conducted to understand the fundamental factor that affects driving behaviour in different settings. In this survey, we began by discussing the primary methodologies used to analyze driving behaviour patterns. Considering the analysis of different publications in this area, we find that the field of driving behaviour analysis is increasing due to the interests of different stakeholders in studying driving behaviours for a safer transportation environment.

We have presented a dynamic methodology for the previous review in analyzing driving patterns to identify driving behaviour. We collected several previous studies and categorized them according to the methodologies used in data analysis. We compared the advantages and limitations of the major papers in this field. We realized the importance of a dynamic survey mechanism that enables research to add new research efforts to existing ones. Fortunately, our dynamic survey methodology will assist researchers in this field in the future to better understand the current contributions to driving behaviour analysis. If an article on driving behaviour is published, the dynamic database automatically adds it to our database data warehouse and makes it available for further analysis. The availability of data on this field is crucial to the success of driving behaviour analyses. We discuss several publicly available data sources that can be used to analyze driving behaviour.

Another important point we wish to highlight is that, after analysis of different models and algorithms proposed by previous researchers, we conclude that the precision of algorithms which in some way implemented the subject of time series in their models is higher than those models that deal only deals with data changes regardless of time. The number of articles that considered this field from a graph perspective has so far been minimal. In this way, the vehicle movement sequences to extract driving behaviour are considered consecutive graphs; the existing pattern in modifying these graphs is classified. The most up-to-date articles in this field show the high accuracy of this way of thinking. Fortunately, in the future, our framework will assist researchers in this field in following and analyze the existing trend in this field from other angles.

# Figures:

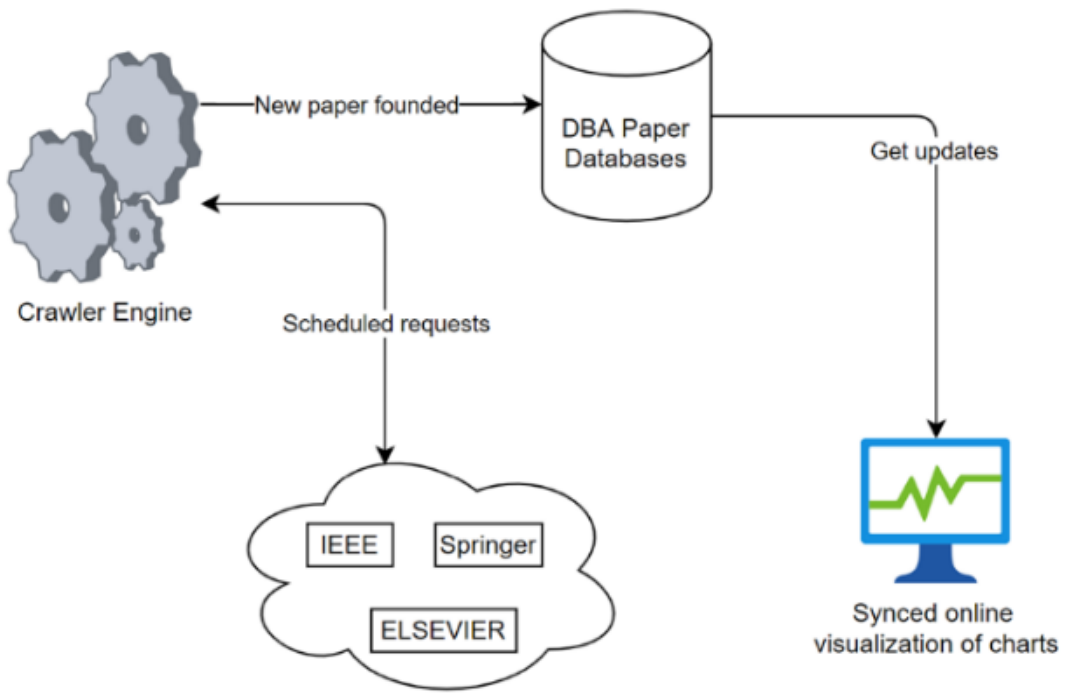

*Figure 1Dynamic survey mechanism*

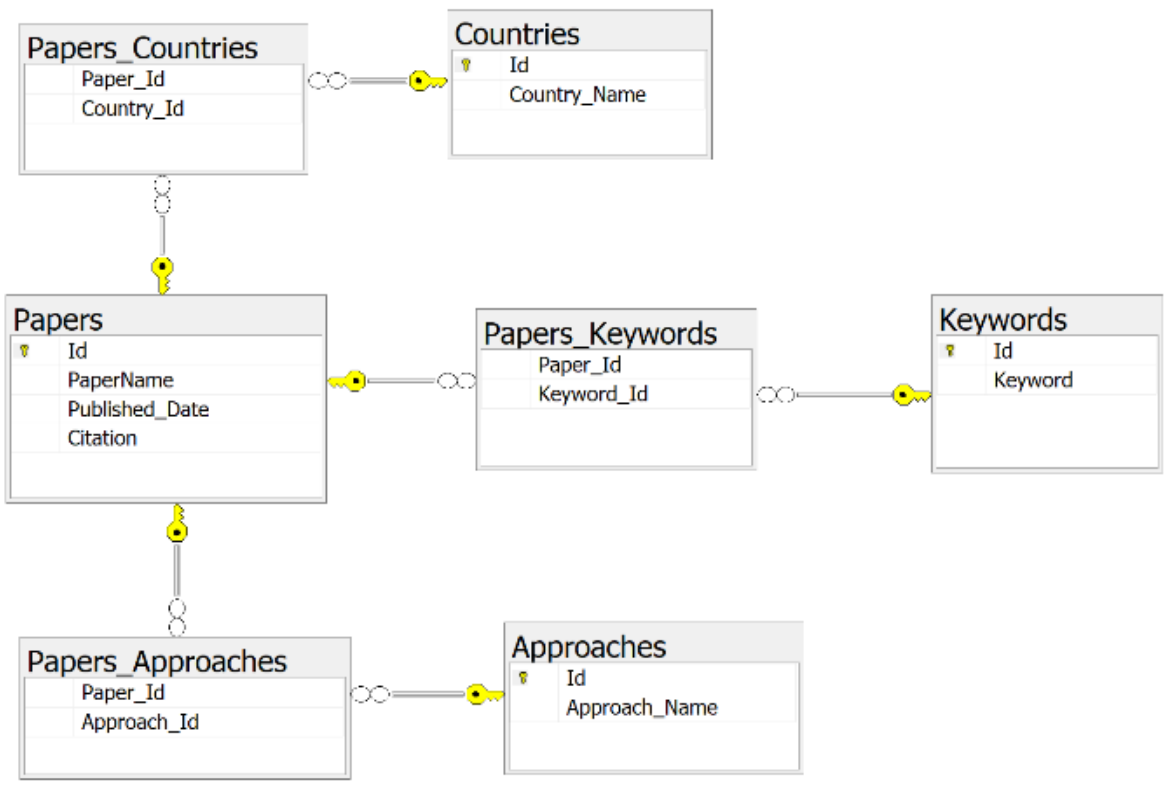

*Figure 2Papers database structure diagram*

| | Country | Paper Count ▾ |
|---|---|---|
| 1. | China | 244 |
| 2. | USA | 101 |
| 3. | Japan | 58 |
| 4. | Australia | 24 |
| 5. | Taiwan | 24 |
| 6. | Korea | 16 |
| 7. | UK | 15 |
| 8. | Singapore | 14 |
| | Grand total | 572 |
| | | 1 - 20 / 20 |

| | Approach | Paper Count ▾ |
|---|---|---|
| 1. | CNN | 115 |
| 2. | LSTM | 107 |
| 3. | Auto Encoder | 91 |
| 4. | RNN | 39 |
| 5. | SOM | 29 |
| 6. | DBN | 21 |
| 7. | DNN | 15 |
| 8. | NK-DNA | 15 |
| | Grand total | 572 |
| | | 1 - 23 / 23 |

| | Keyword | Paper Count ▾ |
|---|---|---|
| 1. | aggressive driving behaviors | 48 |
| 2. | autonomous driving | 36 |
| 3. | driving assistant | 30 |
| 4. | anomaly detection | 21 |
| 5. | deep learning | 18 |
| 6. | deep q-network (dqn) | 18 |
| 7. | cybersecurity | 18 |
| 8. | connected vehicle | 16 |
| | Grand total | 572 |
| | | 1 - 57 / 57 |

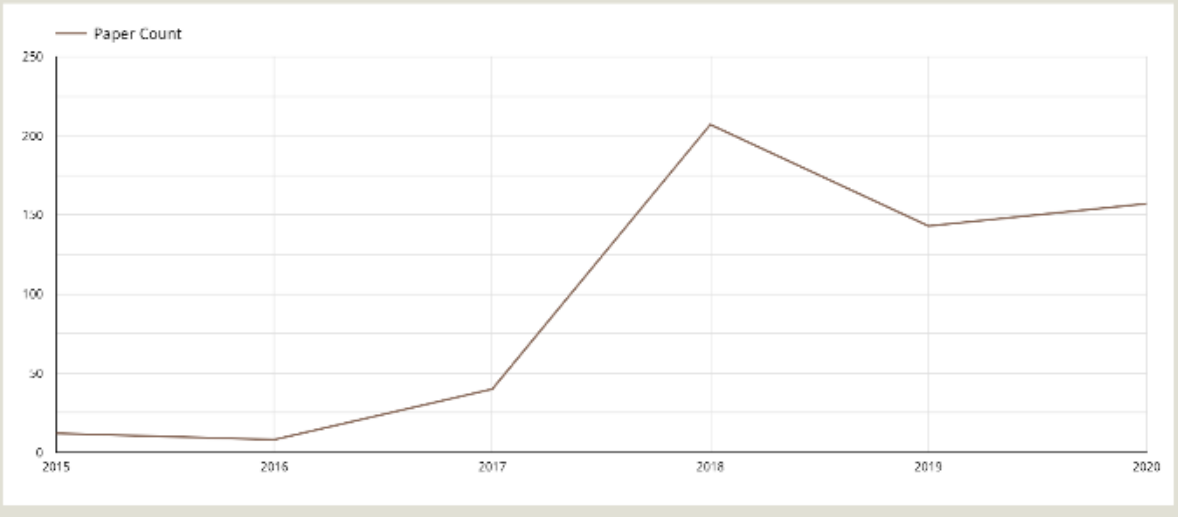

*Figure 3Dynamic sortable charts*

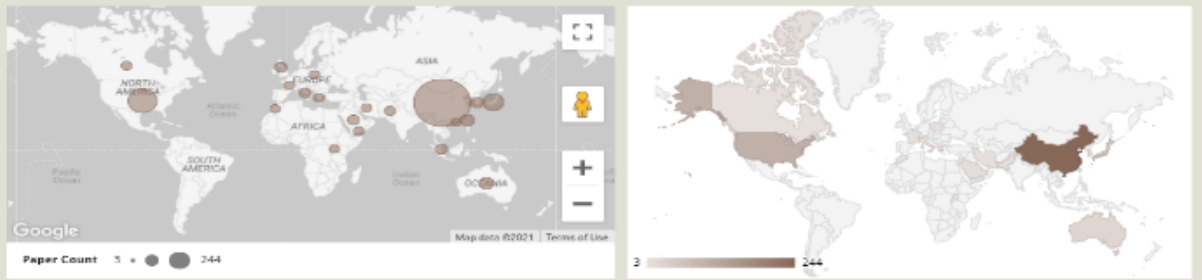

| Country | CNN | LSTM | Auto Encod... | RNN | SOM | DBN | DNN | NK-DNA | null | Grand t... |
|---|---|---|---|---|---|---|---|---|---|---|
| Taiwan | 6 | - | - | - | 6 | - | - | - | - | 24 |
| Korea | 4 | - | 4 | - | 4 | - | - | - | - | 16 |
| UK | - | 6 | 3 | 6 | - | - | - | - | - | 15 |
| Singapore | - | - | 5 | - | 5 | - | - | - | - | 14 |
| Saudi Arabia | 6 | - | - | - | - | - | - | - | 7 | 13 |
| Italy | 6 | - | 5 | - | - | - | - | - | - | 11 |
| Pakistan | - | - | - | - | - | - | - | - | 7 | 8 |
| Greece | - | - | - | - | - | - | - | - | - | 6 |
| Canada | - | 5 | - | - | - | - | - | - | - | 6 |
| Yemen | 6 | - | - | - | - | - | - | - | - | 6 |
| Iran | - | - | - | - | - | - | 5 | - | - | 5 |
| France | - | - | 5 | - | - | - | - | - | - | 5 |
| Poland | - | - | - | - | - | - | - | 5 | - | 5 |
| Macao | - | 4 | - | - | - | - | - | - | - | 4 |
| Grand total | 115 | 107 | 91 | 39 | 29 | 21 | 15 | 15 | 14 | 572 |

*Figure 4Geographical distribution of studies on DBA*

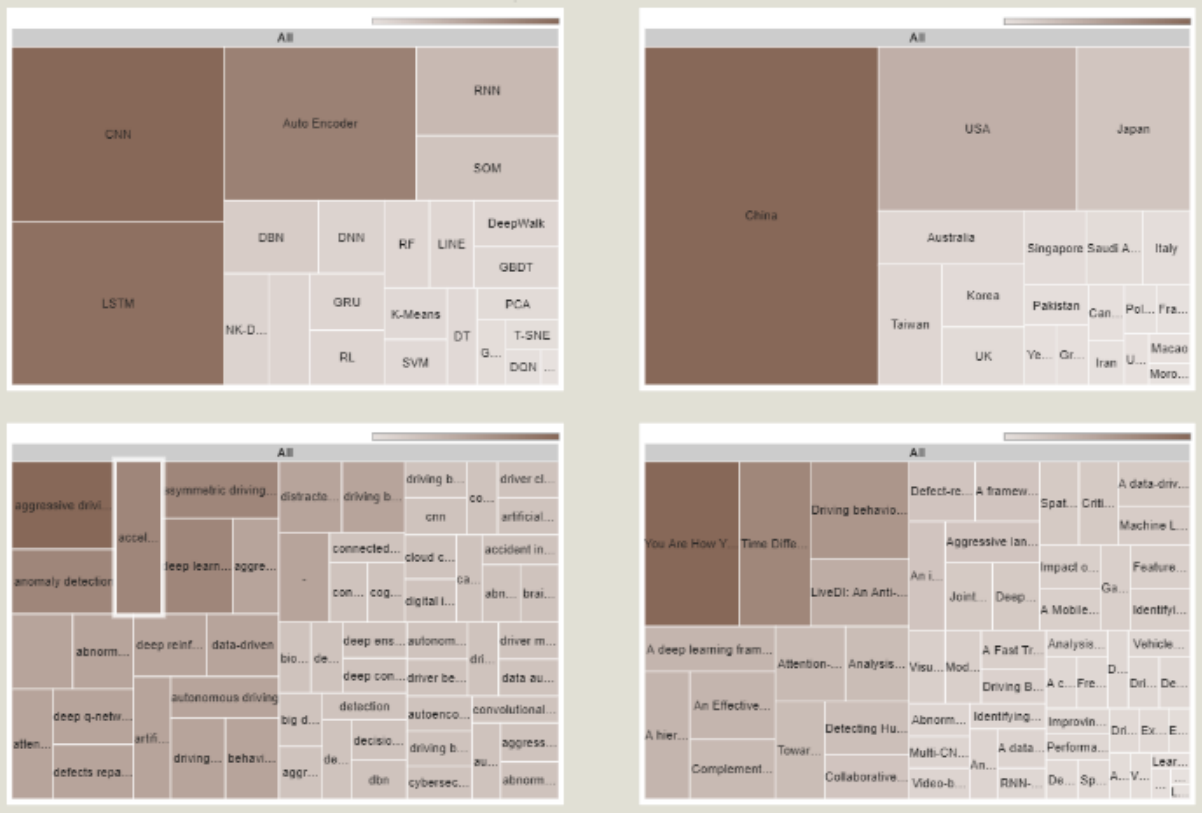

*Figure 5Treemaps of different features in studied papers*

| | Approach_Name | Record Count | Citation ▾ |
|---|---|---|---|
| 1. | LSTM | 107 | 13 |
| 2. | CNN | 115 | 12 |
| 3. | Auto Encoder | 91 | 8 |
| 4. | SOM | 29 | 4 |
| 5. | RNN | 39 | 3 |
| 6. | DBN | 21 | 3 |
| 7. | RF | 12 | 2 |
| 8. | GRU | 13 | 2 |
| 9. | K-Means | 10 | 2 |
| 10. | SVM | 9 | 2 |
| 11. | DNN | 15 | 2 |
| 12. | RL | 13 | 2 |
| 13. | GHSOM | 6 | 1 |
| 14. | DDPG | 2 | 1 |
| 15. | LINE | 12 | 1 |
| 16. | DT | 9 | 1 |
| 17. | T-SNE | 5 | 1 |
| 18. | NK-DNA | 15 | 1 |
| 19. | DQN | 4 | 1 |
| 20. | No data | 14 | 1 |
| 21. | PCA | 8 | 1 |
| 22. | DeepWalk | 12 | 1 |
| 23. | GBDT | 11 | 1 |
| | Grand total | 572 | 21 |
| | | 1 - 23 / 23 | |

*Figure 6Approaches, Counts, and Citation counts*

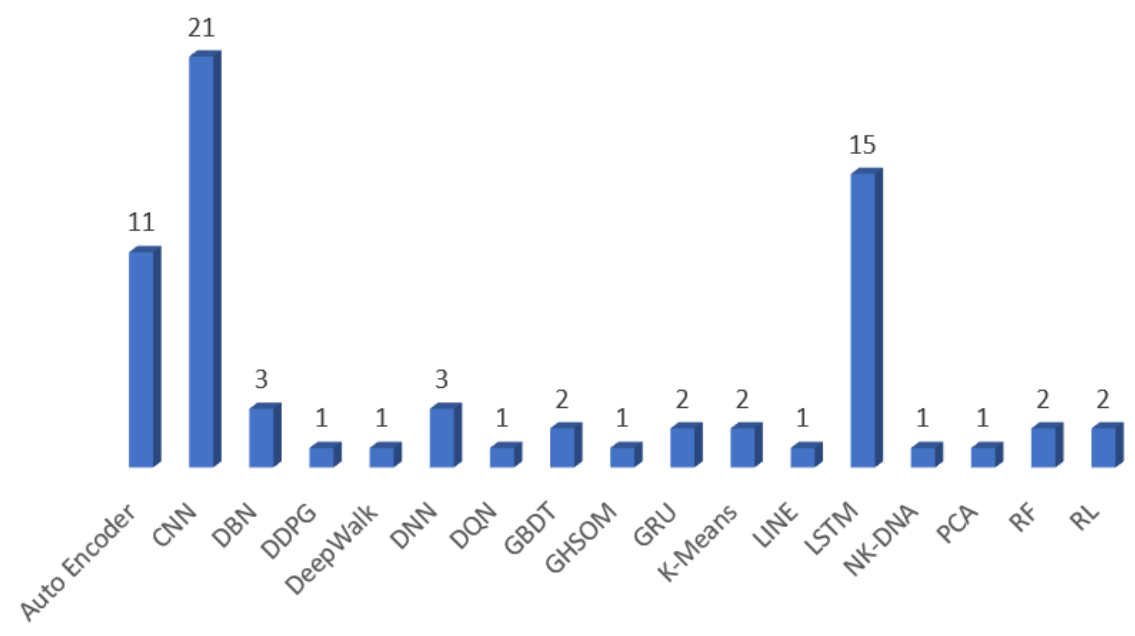

*Figure 7Approaches used counts*

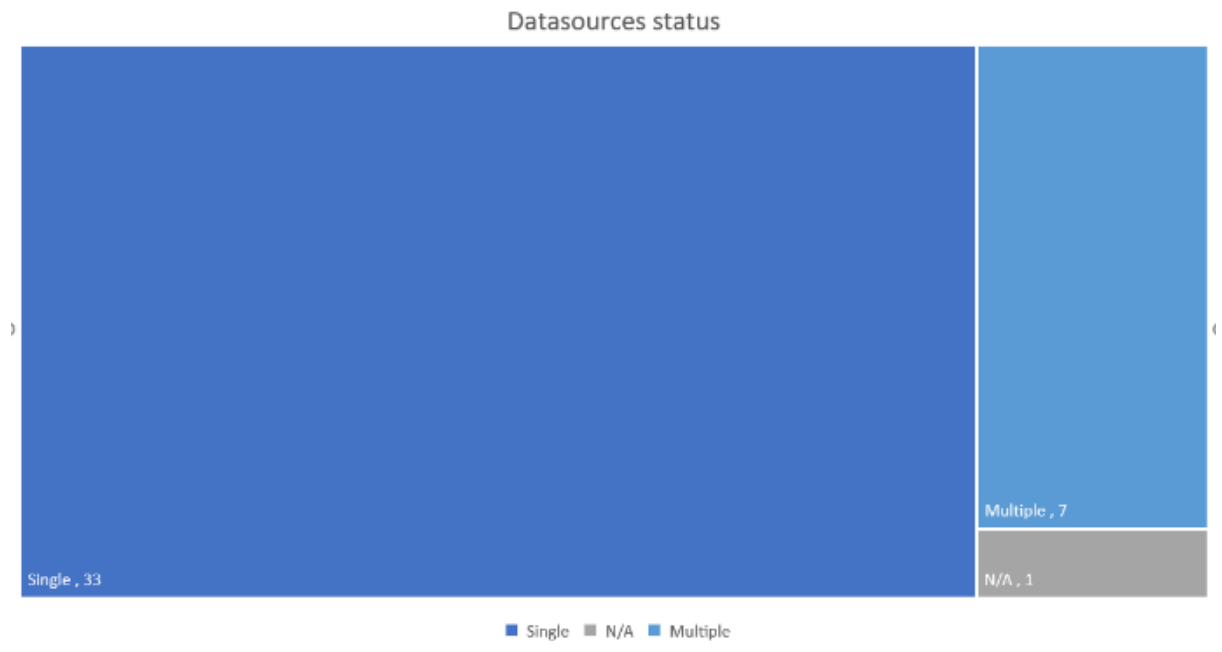

*Figure 10Sources of data status*

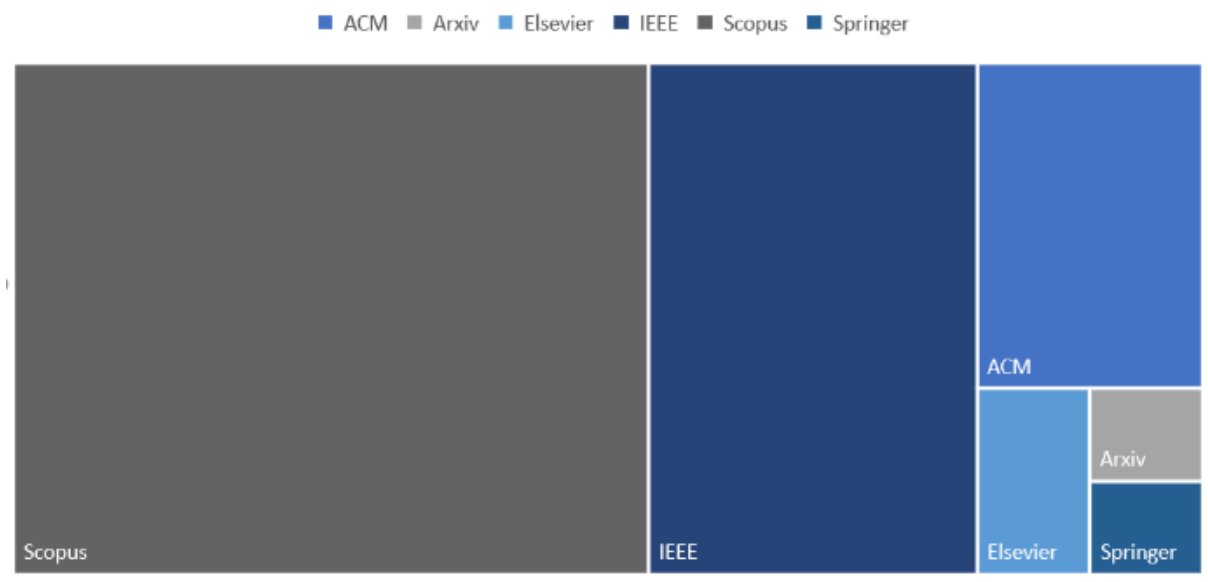

*Figure 8Database's paper counts in DBA field*

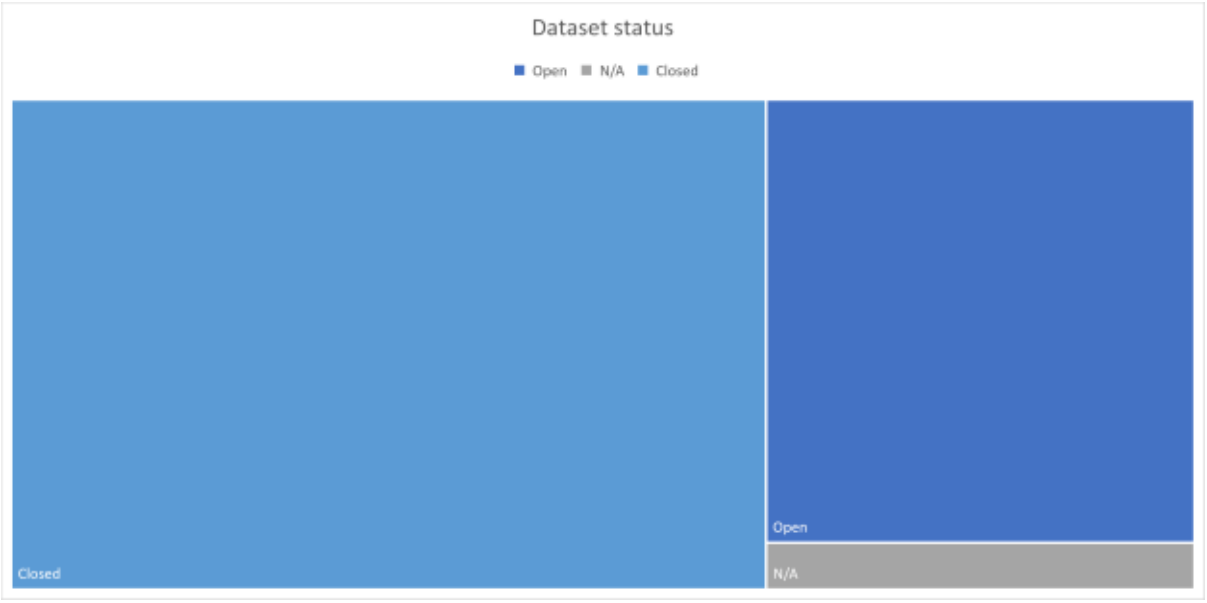

*Figure 9Dataset statuses of reviewed papers*

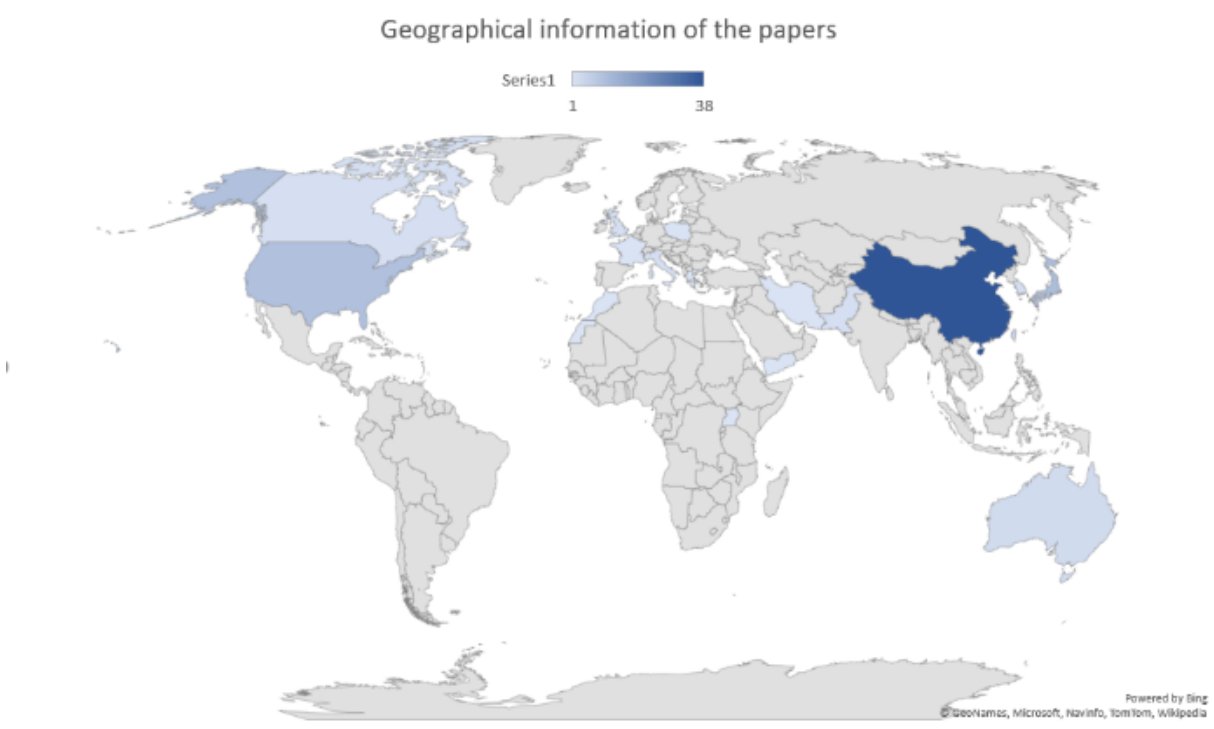

*Figure 11 Geographical distribution of publications*

| Article | Proposed approaches | Citation | Dataset | Summary |
|---|---|---|---|---|
| **[25]** | CNN; Pooling; RNN; | 77 | Kaggle. Driver Telematics Analysis. [27] | For DL-based driving behaviour analysis, GPS data was utilized; extracting important attributes explains behaviour patterns. Human effort and effort both suffered. |
| **[26]** | CNN | 0 | The Didi Chuxing GAIA Initiative to the research community; [39] | The paper proposes dividing drivers into four types: risky, dangerous, safe, and low risk. This paper's main contribution is using CNNs (Convolutional Neural Networks) to process raw trajectories into inputs for the CNNs. After training the CNN network, 77.3% accuracy was achieved. |
| **[40]** | RuLSIF; SOM; Clustering; Deep auto-encoder; | 1 | Kaggle. Driver Telematics Analysis. [27] | The vehicle data recorder was replaced with an internal smartphone sensor. However, unlabeled telematics data limits their application in analyzing driving patterns. Unsupervised learning was used to obtain mobile telematics data. Three significant components included a self-organizing map, a nine-layer deep auto-encoder, and partitive clustering algorithms. |
| **[41]** | Resampling; Normalization ; Stacked-LSTM; | 47 | U-AH Dataset [35] | Recent Stack LSTM Recurrent Neural Networks for classifying driving behaviour; time-series classification was applied using smartphone sensors. The Stacked-LSTM model was validated using the dataset known as UAH-DriveSet. An LSTM stack excelled on the UAHDriveSet. |
| [34] | Abrupt change detection; Sparse auto-encoder; Two-level clustering (SOM; K-means) | 16 | DTG database maintained by the TS (Traffic Safety) | The findings find habits in driving. Three different data analytic analysis methods were used, including abrupt change detection. This model was developed on data from 43 Korean city taxis. The framework can find aggressive driving clusters in large-scale driving records. |
| [36] | Two-level K-Means | 8 | OSeven Telematics, London, United Kingdom | Information about harsh events occurrence, acceleration profile, mobile usage, and speeding was used in this study to detect unsafe driving behaviour. To separate aggressive from nonaggressive trips, an initial clustering was done. A second-level clustering was done to delineate "normal" trips from unsafe ones. |
| **[42]** | Multi-CNN; GBDT predictor | 1 | KITTI dataset [2] | Feature integration is critical, which is why they propose the multi-CNN architecture. A dynamic fixed point compression method is applied in our system; smaller model size and faster speed can be achieved while the accuracy is high; prediction results are a driving score that reflects driver behaviour. |
| **[43]** | maximum entropy inverse reinforcement learning (MEIRL) with automatic feature correction | 22 | N/A | The building block for many LBS applications is trajectory outlier detection. The paper focuses on accurately detecting outliers in-vehicle trajectories. The proposed solution uses a late-stage alarm for a missed outlier trajectory (i.e., the trajectory has not yet reached the destination). |
| **[28]** | Auto-Encoder; Deep Walk; CNN; LINE | 32 | T-Drive trajectory dataset [37] | A GPS trajectory analysis framework (PTARL) was developed. GPS traces were used to track driving operations and driver states. When determining a driver's behaviour, multiview driving state transition graphs were used. A representation learning method for sequence learning from time-varying yet relational state transition graphs were produced. According to the method, both graph-graph dependency and temporal dependency can be handled using a unified optimization framework. |
| **[44]** | Pattern recognition; GHSOM; | 3 | N/A | A pattern recognition process was used to model the driving pattern based on one driver and a fleet's energy consumption. The GHSOM shows that learned driving behaviours can be recognized as the number of driving cycles increases. Additionally, the proposed |
| | | | | framework would enhance driver behaviours and make it easier to design an ADAS. |
| **[45]** | SFG; GRU (an enhancement of LSTM) combine with FCN; | 1 | HACKING AND COUNTERMEASURE RESEARCH LAB. [33] | This paper presents a new technique, LiveDI, which uses driving behaviour to identify drivers. The model uses GRU and FCN to learn long-short term patterns of driving behaviours from drivers. Additionally, training time was increased by implementing the SFG algorithm to identify a time window for analysis. |
| **[46]** | GHSOM (Unsupervised) + SVM (Supervised) | 5 | N/A | A pattern recognition approach is proposed to model the driving pattern, given the consumption of an EV. Gradual drivers' behaviour change is implemented through the GHSOM, and classifiers are implemented with clustered neurons for an online process using SVM. |
| **[47]** | CGARNN-Edge | 9 | N/A | A "pBEAM" platform for personalized driving behaviour modelling is promoted. The driving behaviour model is built on top of "GARNN" which follows dynamic changes in everyday driving. Moving models to the edge improves model performance and robustness. "CGARNN-Edge" is a model tailored to drivers' personal information and preferences as additional conditions. |
| **[48]** | SVM; RF; NN | 4 | N/A | This paper uses ML methods to analyze and predict driving routes, thereby establishing a solid foundation for improving driver behaviour. |
| **[49]** | DNN; LSTM | 12 | Kaggle. Driver Telematics Analysis. [27] | This study analyzes sensor data to identify sematic-level driving behaviour. A large dataset was utilized layer-by-layer for driving maneuvers. The specific maneuver driver ID is helpful in supervised learning of higher-level feature abstraction. This paper proposes a joint histogram feature map to normalize the "Shallow features" for DL. The results show that DNN is suitable for classifying driving maneuvers, with 94% accuracy, whereas LSTM NN has the accuracy of 92% when identifying a specific driver. |
| **[50]** | LSTM; | 9 | N/A | The perceived risk by drivers in different groups on a two-lane road is tested, using a DNN method to summarize environmental features. This training and testing data is for the learning network. Using an LSTM model, risk perception is modelled as a function of traffic conditions and vehicle data. |
| **[51]** | PCA; Fast ICA; KPCA | 89 | N/A | The DSAE method is used to uncover previously hidden driving features for visualization. The DSAE has a method of producing a driving colour map using 3-D hidden feature extraction to RGB colour space. For a driving map, colours are placed in the corresponding map locations. |
| **[52]** | Rep-DRQN; LSTM; | 1 | Simulated data (SUMO) | Traffic flow can be improved by applying reinforcement learning techniques to a traffic control system. First, a microcosmic state representation, which integrates vehicle dynamics, such as lane changing, car-following, and previous phases of a traffic light, is proposed. The red light flooding the system is also incorporated into the action space. A partially LSTM network is used to improve travel experience and efficiency. In practice, parallel sampling is used to speed up training convergence. |
| **[53]** | LSTM; | 3 | HACKING AND COUNTERMEASURE RESEARCH LAB. [33] | The uniqueness of a driver's driving behaviour helps driver profiling and vehicle security (anti-theft systems). This paper analyzes data-driven end-to-end models intended for behaviour identification and examines the principles that underlie the model designs. The real-world driving dataset is employed in cross-validation to test various data-driven DL and machine learning models. |
| **[54]** | CNN; LSTM | 4 | Gathered by themselves | This promising research employs a two-stream CNN approach for video-based driving behaviour recognition. CNN collects motion information by computing optical flow displacement over a few adjacent frames. A spatial-temporal fusion study was conducted to determine behaviour recognition, constructing a 1237-video dataset simulating different driving behaviours to test the model's efficacy. |
| **[55]** | Deep RL | 10 | Simulated data | A decision-making method based on deep reinforcement learning is proposed for connected vehicles in complex traffic scenarios. The |
| | | | | model has three primary components: a data preprocessor that transforms hybrid data into a grid matrix data format; a two-stream deep neural network that extracts the hidden features; and a deep reinforcement learning network that learns the optimal policy. Additionally, a simulation environment is built to train and test the proposed method. The results show that the model can learn the best overall driving policy, such as driving fast through diverse traffic without unnecessary lane changes. |
| **[56]** | DedistractedNet; CNN; | 4 | Gathered by themselves | DedistractedNet was built to identify distracted driving behaviours from an image. DedistractedNet uses neural networks to identify driving behaviour features without onboard diagnostics or sensors. The experiments show that the DedistractedNet performs better than the other baseline CNN methods. |
| **[57]** | MV-CNN + Data Augmentation | 14 | Gathered by themselves | This paper introduces a driver behaviour recognition system utilizing a six-axis motion processor. DL learns from onboard sensor data. A new algorithm is proposed (called MV-CNN) that includes the multi-axis weighted fusion algorithm, background noise fusion algorithm, and random cropping algorithm. Following the CNN model, a new model, called MV-CNN, was developed. |
| **[58]** | Stacked Auto-Encoders; | 0 | Simulated data | DL is the proposed method of classifying individual drivers (stacked autoencoders). Sensor signals from a driving simulator were used to assess drivers' driving skills. The maximum driving skill recognition rate was 98.1 percent, and the recognition rate was also increased in this research. |
| **[59]** | DBN | 6 | Gathered by themselves | A deep belief network (DBN) was used to build the learning model, and training data was collected from real-world road drivers. Using the model, the front wheel's steering angle and the vehicle's speed are predicted. Prediction results show that DBN has higher accuracy and adapts to different driving scenarios with fewer modifications. |
| **[60]** | Reviewing all approaches; | 4 | N/A | HIDB is categorized into two major categories: Driver Distraction (DD), Driver Fatigue (DF), or Drowsiness (DFD). Aggressive Driving (ADB) is also discussed. ADB is a wide range of driving styles that have significant consequences. DD, DFD, and ADB are affected by the experience, age, and gender or illness. |
| **[61]** | Deep Auto Encoders; | 4 | Gathered by themselves | A new approach to proactive driving using human experts and autonomous agents is introduced. DL methods extracted latent features. Velocity profiles were created to provide an autonomous driving agent with human-like driving skills. Being proactive was shown to help avoid unnecessary jerkiness. |
| **[62]** | Spectral Clustering Algorithm; LSTM; | 19 | Gathered by themselves | Researchers identified a macroscopic relationship between driving behaviour and fuel consumption in the natural driving process using unsupervised spectral clustering. Additionally, dynamic information was acquired from the driving environment and driving data to link different driving behaviours to fuel consumption features to give computers the ability to recognize environments. The vehicle's operating signal data was used to provide the training data for the DL network. Fuel consumption feature distribution was based on roadway data and historical driving data. |
| **[63]** | LSTM; | 0 | The Warrigal dataset [38] | The researchers have developed (1) a novel feature extraction method for raw CAN bus data;(2) a novel boosting method for driving behaviour classification (safe or unsafe) that combines advantages of DL and shallow learning methods; and (3) a first-of-its-kind public transportation industry evaluation using real-world data to ensure accurate labels from industry experts. |
| **[64]** | Deep Q-Network | 3 | Gathered by themselves | This L-HMC combines deep reinforcement learning with collision avoidance capacity. An improved DQN method is used to learn the best driving policy for pedestrian collision avoidance. The findings demonstrate that the deep reinforcement learning-based method can rapidly learn an effective pedestrian collision avoidance driving policy. Meanwhile, L-HMC uses flexible policies to avoid pedestrian collisions in typical scenarios, improving overall driving safety. |
| **[65]** | Stacked Auto-Encoders; | 3 | Gathered by themselves | This novel DL-based model is built for abnormal driving detection. A stacked sparse autoencoder enables learning driving behaviour features. Training is layer-wise, greedy. This is the first time researchers have used DL to build representations of driving features using autoencoders. The algorithm is also denoised with an algorithm, making it more stable. Dropout is commonly used in training to reduce overfitting. The proposed system has better performance for finding abnormal driving. |
| **[66]** | LSVDNN | 0 | Gathered by themselves | Using the designed model, the output used for controlling the vehicle is obtained. The learning and validation approach for self-driving vehicles (LSV-DNN) is outlined, and a convolutional network based on vehicle cameras and computer data is developed. Obstacle detection is carried out with the best accuracy and speed using the Yolo algorithm version 3. |
| **[67]** | DCNN | 2 | https://github.com/abdugumaei/ADBs-Dataset | A real-time detection system is proposed, utilizing bio-signals and a deep CNN model, which incorporates edge and cloud technologies. The system contains three distinct modules: vehicle edge devices, cloud platforms, and monitoring environments, all of which are linked via a telecommunication network. Processed bio-signal datasets are employed to test the proposed DCNN model. The dataset was collected using a different time window and time step than the bio-signal datasets. |
| **[68]** | Denoising Stacked Autoencoder (SDAE) | 4 | Gathered by themselves | To develop a graphical representation of driver behaviour and the road environment, an improved DL model is proposed in this study. A Denoising Stacked Autoencoder (SDAE) is proposed to provide output layers in RGB colours. The dataset was collected from an in-vehicle GPS tracking device on an experimental driving test. Using graphics, the method efficiently identifies simple driving behaviours and other events encountered along the path. |
| **[69]** | Deep Deterministic Policy Gradient (DDPG); | 0 | Simulated data (TORCS) | Adaptive driving behaviour for simulated cars is proposed using continuous control deep reinforcement learning. The DDPG delivers smooth driving maneuvers in simulated environments. Recurrent Deterministic Policy Gradients were used to encode time (or Recurrent DDAGs). A trained agent adapts to traffic velocity. |
| **[70]** | DBN; LSTM; | 42 | Simulated data (NGSIM, 2006) | This paper uses data-driven LC modelling using DL. To better model the LC process, Deep Belief Network (DBN) and Long Short-Term Memory (LSTM) neural networks are employed (LCI). The NGSIM project's empirical LC data is used for training and testing the proposed DBN-based and LSTM-based LCI models. |
| **[71]** | DBN-FS (Fuzzy sets) | 1 | Gathered by themselves | The feature matrix contains data on nearly 2,000 lane-change videos. Also, vehicle information is obtained based on license plates. A state-of-the-art DBN DL algorithm creates a lane-change behaviour model that incorporates relevant vehicle, driver, and driving variables. The model's superior accuracy, feasibility, and concreteness are verified by comparison with other common models. |
| **[72]** | CNN | 0 | Simulated data (OpenAI) | DL-based learning techniques are proposed in this paper, which can be applied in various driving scenarios. The proposed method is tested for effectiveness and efficiency, and the proposed methods are shown to outperform other ML methods. |
| **[73]** | LTSM-FCN | 5 | U-AH Dataset [35] | The solution proposed in this paper is based on a Long Short-Term Memory Fully Convolutional Network (LTSM-FCN) to identify driving sessions that include aggressive behaviour. The problem is formulated as a time series classification, and the validity of the approach is tested on the UAH-DriveSet, a dataset that provides naturalistic driving data collected from smartphones via a driving monitoring application. |
| **[74]** | CNN | 18 | SEU-DRIVING / KAGGLE-DRIVING | A DL method for classifying driving behaviour in a single image is investigated in this paper. The classification of driving behaviour is a multi-class problem. In two ways, the research team discovered a solution to this problem: Extract multi-scale features using multi-stream CNN was extracted and then combined into a final decision for driving behaviour recognition. |
| **[75]** | gradient boosted model with grid search; | 1 | https://insight.shrp2nds.us | The study included multiple factors to produce an interpretable model for accident occurrence, given road conditions and driver behaviour. Seven thousand seven hundred trips were studied using four ML and DL techniques. Accident prediction was achieved by a gradient boosted model. Predictive factors were shown to be primary behaviour, pre-incident maneuvers, and secondary task duration. |
| **[76]** | Feature selection with NN | 1 | Gathered by themselves | Feature extraction and a DL model are suggested to detect abnormal driving behaviour. This method was developed based on bin variation calculation and subsequent feature generation. Variance similarity was used to expand the subset. Variance data from data segments with specific driving behaviour class definitions revealed the connection. Driving behaviours included weaving, sudden braking, and everyday driving. |
| **[32]** | CNN; LSTM; GRU | 22 | HACKING AND COUNTERMEASURE RESEARCH LAB. [33] | A unified end-to-end DL framework based on convolutional neural networks and recurrent neural networks is proposed for time series CAN-BUS sensor data. This method is capable of learning driving and temporal information without prior knowledge. The method can access rich feature representations of driving behaviours from multi-sensor data. |
| **[77]** | NCAE; MC-CNN; | 1 | HACKING AND COUNTERMEASURE RESEARCH LAB. [33] | The researchers first propose utilizing an unsupervised three-layer nonnegativity-constrained autoencoder to search for the sliding window's optimal size and then build a deep nonnegativity-constrained autoencoder network to complete driver identification. Their method can search for optimal window size and save many data compared to conventional sparse autoencoder, dropout-autoencoder, random tree, and random forest algorithms. Also, their technique helps classifiers distinguish the differing class boundaries. Finally, their method helps increase the prediction time and reduce model overfitting. |
| **[78]** | Decision Tree; Random Forest; | 13 | https://insight.shrp2nds.us | ML is suggested to identify secondary tasks drivers use while driving. First, drivers' distraction is found, and second, unique kinds of distractions are discovered. Nine classification methodologies are utilized to identify three secondary tasks (hand-held cellphone calling, cellphone texting, and interaction with an adjacent passenger). The models use five driving behaviour parameters (including standard deviations) as inputs. The paper's findings show that using a proposed methodology for characterizing drivers' involvement in secondary tasks (like texting) helps drivers identify driving hazards and alert them to problems on the road. |
| **[79]** | CNN; RNN | 9 | Gathered by themselves | This paper proposes a DL framework for behaviour extraction. The machine used for their method models temporal features captures salient structure features and fuses CNN and RNN with an attention unit. Gathering a driving behaviour dataset also takes into consideration gravity's effect. Device-independent sensor data is collected. The preferred sensor information is furnished by this method. |
| **[80]** | CNN; FC; LSTM; | 0 | Gathered by themselves | A driver's eyes are the primary source of information while they are driving. Data show that drivers' gazes precede and correlate with driving maneuvers. Thus, GazMon is designed to detect and predict driving maneuvers. The GazMon facial analysis uses facial landmarks, including facial features and head posture, to evaluate the effects. Their GazMon outperforms the competing products in predicting and reducing distracting behaviours. It is easy to customize and will work with existing smartphones. |
| **[81]** | DenseNet | 6 | https://www.kaggle.com/c/state-farm-distracted-driver-detection/data | This paper stresses D fusion techniques. For the first time, three novel DL-based fusion models for abnormal driving behaviour detection are proposed. WGD network, WGRD network, and AWGRD network are three DL-based fusion models equivalent to their functional characteristics. The actual model structure of WGD is modelled using DenseNet. |
| **[82]** | Autoencoders | 10 | Gathered by themselves | This paper studies encoding, clustering, and modelling driver behaviours to build an autonomous vehicle agent. A typical |
| | | | | Japanese suburban area driver provided driving speed, braking, steering, and acceleration data. A fully Connected Deep Autoencoder was used to long datasets of consecutive measurements to collect driving data for clustering purposes. Data were modelled and validated in a ROS car simulator. |
| **[83]** | t-SNE; CVAE; | 2 | Gathered by themselves | This paper offers a DL approach to vehicle driving styles. The neural network groups short behavioural segments into a latent space. The driving dataset had 59 drivers on a highway. Embedded driving behaviour data were clustered into clusters using a topological map. Elements exhibit probabilistic distributions that compactly describe driving episodes. |
| **[84]** | DCNN | 1 | Gathered by themselves | This paper focuses on the end-to-end technique that emulates human drivers' decisions, such as steering angle, acceleration, and deceleration. Ignoring previous states is investigated by comparing predicted accuracy and variation, using data collected in a simulation study. |
| **[85]** | DNNR-Ensemble | 0 | N/A | By using PAYD, insurance carriers can avoid unjust and inefficient policies. The authors propose a PAYD method that incorporates user behaviour factors from multiple dimensions. This first dimensionally divides all factors. Treating each dimension of the factors separately improves the model's efficiency. DNNRegressor is used to make each classification dimension weak. The DNNRegressor classifier yields the final output. |
| **[86]** | LSTM; | 51 | Next Generation Simulation (NGSIM) | This paper proposes an LSTM NN-based car-following (CF) model to capture realistic traffic flow characteristics. The proposed CF model is calibrated and validated using NGSIM data. Three driving-related characteristics are investigated: hysteresis, discrete driving, and intensity difference. The simulation results show the CF model's good traffic flow features reproduction. |
| **[87]** | R-CNN | 1 | Gathered by themselves | In this paper, vehicle-mounted camera-based driver detection that utilizes Faster R-CNN is proposed. First, a residual structure is added to the ZF network. BN replaces LRN, which increases parameter stability and accelerates network convergence. |
| **[88]** | NK-DNA | 6 | Gathered by themselves | A new security model using driving data and the neural knowledge DNA is proposed in this paper. A novel knowledge representation method helps computers discover, store, reuse, improve, and share knowledge. |
| **[89]** | Autoencoder and Self-organized Maps (AESOM) | 21 | Shenzhen Urban Transport Planning Center, Shenzhen, China | A hybrid unsupervised DL model for modelling driving behaviour and risky patterns was developed for this paper. The extraction method uses Autoencoders and Self-Organized Maps (AESOM). |
| **[90]** | deep convolutional neural network; LSTM; | 6 | Gathered by themselves | Because many traffic accidents occur at intersections due to unsafe driving behaviours, this paper presents a smartphone-based system for analyzing driving behaviour at intersections. A deep convolutional neural network-based model is proposed to detect traffic lights, crosswalks, and stop lines. The LSTM-based model estimates vehicle speed using an accelerometer and gyroscope embedded in the smartphone. |
| **[91]** | DSAE | 4 | N/A | A driving behaviour time series is assumed to be generated from a single-dimensional dataset that everyone has access to. Sensor time-series data is faulty because of a sensor failure. Another essential function is to limit the negative impact when extracting low-dimensional time-series data. Using a DSAE, low-dimensional time-series data is extracted. |